\documentclass{article}

\usepackage[nonatbib, preprint]{neurips_2020}

\usepackage[utf8]{inputenc} 
\usepackage[T1]{fontenc}    
\usepackage{url}            
\usepackage{booktabs}       
\usepackage{amsfonts}       
\usepackage{nicefrac}       
\usepackage{microtype}      

\usepackage[numbers]{natbib}
\usepackage{graphicx}
\usepackage{amsmath}
\usepackage{color}
\usepackage{caption} 
\usepackage{url}

\usepackage[breaklinks]{hyperref}
\usepackage{breakurl}

\DeclareMathOperator*{\argmin}{arg\,min}

\newif\ifcomments
\commentstrue

\title{
Fairness and Robustness in Invariant Learning: 
\\ A Case Study in Toxicity Classification
} 

\author{
  Robert Adragna \\
  Department of Computer Science\\
  University of Toronto\\
  \texttt{robert.adragna@mail.utoronto.ca} \\
   \And
   Elliot Creager \\
   Deparmtent of Computer Science \\
   University of Toronto \\
   \texttt{creager@cs.toronto.edu} \\
   \AND
   David Madras \\
   Department of Computer Science \\
   University of Toronto \\
   \texttt{madras@cs.toronto.edu} \\
   \And
   Richard Zemel \\
   Department of Computer Science \\
   University of Toronto \\
   \texttt{zemel@cs.toronto.edu} \\
}

\begin{document}

\maketitle

\begin{abstract}

   Robustness is of central importance in machine learning and has given rise to the fields of domain generalization and invariant learning, which are concerned with improving performance on a test distribution distinct from but related to the training distribution. In light of recent work suggesting an intimate connection between fairness and robustness, we investigate whether algorithms from robust ML can be used to improve the fairness of classifiers that are trained on biased data and tested on unbiased data. We apply Invariant Risk Minimization (IRM), a domain generalization algorithm that employs a causal discovery inspired method to find robust predictors, to the task of fairly predicting the toxicity of internet comments. We show that IRM achieves better out-of-distribution accuracy and fairness than Empirical Risk Minimization (ERM) methods, and analyze both the difficulties that arise when applying IRM in practice and the conditions under which IRM will likely be effective in this scenario. We hope that this work will inspire further studies of how robust machine learning methods relate to algorithmic fairness.
   
\end{abstract}

\section{Introduction}

Machine learning systems struggle to learn predictors that are robust to distribution shift. When tested on i.i.d data drawn from the training distribution these systems can achieve nearly perfect accuracy, even when regularized to prevent over-fitting. However, performance can degrade to below-chance accuracy when the testing and training distributions are even slightly different \citep{goodfellow2015explaining,geirhos2020shortcut}. The field of Domain Generalization (DG) \cite{ganin2016domainadversarial, li2018domain} addresses this challenge by proposing robust methods that ensure good test performance on distributions that are different from but systematically related to the training distribution \citep{mansour2019robust}. Invariant Risk Minimization (IRM) \citep{arjovsky2019invariant} is a one of several recently successful approaches to Multi-Source Domain Generalization \citep{zhao2020multisource} which encourages models to learn predictors with invariant performance across different ``domains'', or ``environments'' \citep{peters2015causal, heinzedeml2018invariant}. Given $n$ different training environments, these models extract a set of predictors from the feature space such that the conditional distribution of the outcomes given the predictors is invariant  across all training environments. These predictors can consequently generalize well to all test out-of-distribution (OOD) environments which share this same invariance. Building on work in philosophy which characterizes causation as invariance \citep{cartwright2003two, mitchell2000dimensions}, existing invariance-based DG methods have been interpreted as a weak form of causal discovery whose returned predictors are the causal factors underlying the phenomena we wish to predict. 

Fairness can be often characterized by robustness to changes in ``sensitive attributes'' \cite{arjovsky2019invariant}, especially in the context of toxicity classification.      Consider an automated moderation system used by an online news platform to determine which comments on a news article are toxic to online discourse and should be censored. The performance of a fair system should not be affected by characteristics such as whether the comment is about issues related to race, gender or other politically sensitive topics. Alternative definitions of distributive fairness differ in how the system's predictions should be invariant to changes in the sensitive attribute. For instance, statistical definitions such as Demographic Parity require that some conditional distribution of predictions given the sensitive attribute are invariant to the sensitive attribute \citep{hardt2016equality}, and causal definitions such as counterfactual fairness \citep{kusner2018counterfactual} require that every individual's prediction is invariant to counterfactual changes in that individual's sensitive attribute. There are a number of ethical and legal criticisms to be levied against systems that predict based on sensitive group membership \citep{barocas2016big}. Moreover, over-reliance on sensitive information could decrease robustness when the predictive performance of this information spuriously depends on the environmental context in which it is employed. Discussion about non-caucasian racial identities, for example, may be highly predictive of comment toxicity on white supremacist internet forums whose members routinely make discriminatory remarks about ethnic minorities. However, on other internet forums that are more welcoming of diversity the association between racial identity mention and toxicity would likely be far weaker. This brittleness of sensitive information has been identified as a key challenge that Perspective API, a Google-backed internet comment toxicity classifier, faced during implementing in real-world contexts \citep{borkan2019nuanced, jigsawmedium}, and has also been observed to cause bias in sentiment analysis \citep{kiritchenko2018examining} and facial detection \citep{buolamwini18a} tasks. Fair models, then, can perhaps be constructed by learning predictors whose performance remains invariant across a variety of different environments. 

In this work, we empirically demonstrate that Domain Generalization can used to build fair machine learning systems by constructing models that are invariant to spurious correlations involving the sensitive attribute. Specifically, we assess 
the performance of IRM on a fair internet comment toxicity classification task derived from the Civil Comments Dataset. In this task, the model must generalize from biased training environments exhibiting a strong but spurious correlation between mention of a particular demographic identity and toxicity to a test environment in which this correlation is reversed. \footnote{\url{https://github.com/adragnar/irm-toxicity-classification}}

Our contributions are as follows: 
\begin{itemize}
    \item We find that IRM outperforms ERM in both generalization accuracy and group fairness on this dataset,
    highlighting the alignment of robustness and fairness for the type of data bias studied;
    \item We examine the reason for this performance improvement via ablation studies: while sensitive information is marginally predictive, IRM exploits variation in this predictive power across environments to find an invariant (and thus more fair) solution;
    \item Examination of the IRM solution reveals that comment length is used as an invariant feature.
    While predicting from comment length may be more fair than predicting from sensitive group membership, IRM has still not found a truly ``causal'' feature. This highlights both a key limitation of IRM approach of treating invariance as a proxy for causality, and the importance of environment diversity as an inductive bias for this method.
\end{itemize}

\section{Background \& related work}
\subsection{Invariant Risk Minimization}
We consider a set of environments $e \in \mathcal{E}$, each corresponding to a distinct data distribution $D^e$ over some input and label space $\mathbb{X} \times \mathbb{Y}$, and associated risk functions $R^e := \mathbf{E}_{(X^{e}, Y{e}) \sim D^e}[L(f(X^e), Y^e)]$ for arbitrary convex and differentiable loss function $L$. The set of environments is partitioned into multiple training environments $\mathcal{E}_{tr} \subset \mathcal{E}$ and a testing environment $\mathcal{E}_{test} \in \mathcal{E}$ which is inaccessible to the model during training. 

IRM learns a compositional model $w \circ \Phi$, where $\Phi: \mathbb{X} \rightarrow \mathbb{H}$ is a representation function that extracts features from the input space and $w: \mathbb{H} \rightarrow \mathbb{Y}$ is a classifier from these features to outputs. $\Phi$ is said to elicit an invariant predictor $w \circ \Phi$ across all environments $\mathcal{E}$ if there exists some $w$ that is simultaneously optimal for environments. IRM aims to learn an invariant predictor across training environments $\mathcal{E}_{tr} \subset \mathcal{E}$ that minimizes the total risk incurred across all environments. This is formalized through the following bi-level optimization problem:

\begin{align}
     & \min_{\Phi}\sum_{e \in \mathcal{E}_{tr}} R^{e}( w \circ \Phi) \nonumber \\
     \text{s.t.} \ \ \ \ & w \in \argmin\limits_{\hat{w}: H \rightarrow Y} R^{e}(\hat{w} \circ \Phi) \ \forall \ e \in \mathcal{E}_{tr} \label{eq:2}
\end{align}

Given the intractability of this challenging problem, \citep{arjovsky2019invariant} proposes a more practical variant of IRM that approximates Equation \ref{eq:2} - IRMv1

\begin{align}
    \max\limits_{\Phi} \sum_{e \in \mathcal{E}_{tr}} R^{e}(\Phi) + \lambda ||\nabla_{w|w=1.0} R^e(w \circ \Phi)||^2
    \label{eq:irmv1}
\end{align}

Where $\Phi: \mathbb{X} \rightarrow \mathbb{R}^k$ for input space dimension $k$, $w = 1.0$ is a scalar fixed ``dummy classifier'' and $\lambda$ is a hyperparameter balancing the model's predictive performance and its invariance. Equation \ref{eq:irmv1} is equivalent to Empirical Risk Minimization (ERM) over the given loss function with the addition of an invariance penalty for each environment that controls the closeness of $w$ to the true optimal classifier $\tilde{w}^{e}$ for that environment. 

Recent work \citep{arjovsky2019invariant} empirically validates IRM on the Colored MNIST (CMNIST) experiment setup. The digit in each MNIST image is assigned one of two colours based on its identity, and the images are separated into different environments. Then, digit colour is reversed for a different proportion of each environment's images such that there is a strong but spurious correlation between colour and digit identity in the training environments that reverses in the test environment. The task is to predict digit identity on the testing environment after training on the training environments. While ERM models learn to predict identity using digit colour and consequently achieve only 17\% accuracy on the testing environment, IRM learns to predict from the weaker but invariant correlation between digit shape and identity, and consequently achieves a much higher performance of 66.9\% accuracy on the test environment.

The authors of \citep{choe2019empirical} apply IRM in a natural language context analogous to CMNIST. They consider a binary sentiment classification task derived from the SST-2 sentiment analysis dataset in which a strong but spurious correlation between text sentiment and particular forms of punctuation is induced in the training environments that reverses in the testing environment. Our experimental setup builds on this work, but presents a more challenging prediction task for the following two reasons:

\begin{itemize}
    \item In the SST-2 experiments the spurious covariate (punctuation) is easily inferable from the text data given their use of a bag-of-words model \citep{choe2019empirical}, whereas in our experiments the spurious covariate (demographic identity) is not as directly inferable from the data. Annotators were asked to assign identities to comments based on their interpretations of the comment content, which may be presented using a wide variety of different word choices and sentence structures \citep{borkan2019nuanced}.
    
    \item In the SST-2 experiments noise is added to the binary labels by corrupting them with a probability of 25\%. We do not add label noise to our experiments, as toxicity labels are known to be quite noisy by themselves \citep{walker-etal-2012-corpus}. However, the noise present in our labels is unlikely to be uniformly applied to each sample. Toxicity annotations of social media text have been found to be biased against sensitive demographics such that it is more likely for comments involving these demographics to be toxic \citep{sap-etal-2019-risk}. This phenomenon may affect the invariance properties of toxicity across different environments, making it more challenging for IRM to recognize the invariant correlation between given features of the comment and toxicity. 
\end{itemize}
  
\subsection{Fairness in machine learning}
Consider a binary classification problem with data $X \in \mathbb{R}^{n}$, targets $Y \in \{0, 1\}$ and sensitive attributes $A \in \{0, 1\}$. The goal of fair prediction is to learn a model $f \colon X \mapsto Y$ whose predictions $\hat{Y} \in \{0, 1\}$ are maximally close to $Y$ while being fair with respect to $A$. 

One common class of fairness definitions involve statistical parity -- identical distributions of predictions -- with respect to some sensitive attribute \citep{dwork2011fairness}. Two common statistical definitions of fairness are Demographic Parity, which enforces that the same proportion of positive predictions are allocated to each sensitive attribute demographic, and Equalized Odds, which enforces equal false positive and false negative rates across each sensitive attribute demographic \cite{hardt2016equality}. For each fairness definition, we define the classifier's distance from satisfying that definition as follows:

Demographic Parity Distance:
\begin{align}
     \Delta_{DP} = | P(\hat{Y}=1|A=0) - P(\hat{Y}=1|A=1)| \label{eqn:dp}
\end{align}
Equalized Odds Distance:
\begin{align}
    \Delta_{EP} &= \frac{1}{2}|P(\hat{Y}=1|A=0, Y=0) - P(\hat{Y}=1|A=1, Y=0)| \nonumber\\
    & + \frac{1}{2}|P(\hat{Y}=0|A=0, Y=1) - P(\hat{Y}=0|A=1, Y=1)|\label{eqn:eo}    
\end{align}

If $\Delta_{DP}$ or $\Delta_{EO}$ is zero, the classifier is perfectly fair with respect to the associated definition. Conversely, if  $\Delta_{DP}$ or $\Delta_{EO}$ is one, the classifier is maximally unfair with respect to the associated definition.

\subsection{Fairness in toxic comment detection}
Work related to fairness in toxicity classification can be divided into two different approaches. One approach focuses on characterizing the potential for bias in toxicity classification systems at all stages of data processing pipeline. Bias against particular demographic groups has been observed in annotations of toxicity classification datasets \citep{sap-etal-2019-risk}, pre-trained word embeddings \citep{bolukbasi2016man} and language models \citep{tan2019assessing}. As a result, the predictions of toxicity classification models trained without considering these factors have been demonstrated to be unfair \citep{borkan2019nuanced}. Given these considerations, \citep{mitchell2019model} proposes a framework for documenting the fairness properties of pre-trained models to prevent their use in contexts where they are known to predict unfairly. Another approach to fairness in toxicity classification focuses on developing methods to mitigate bias in these systems. Several approaches have been proposed to debias word embeddings \citep{ravfogel2020null, sweeny2020reducing}, and various methods have been proposed that directly target the outcomes of toxicity classification systems: \citep{vaidya2019empirical} uses a multi-task learning model with an attention layer that jointly learn demographic information and toxicity in order to reduce bias, whereas \citep{garg2019counterfactual} takes a causal approach that imposes a fairness penalty on models that predict different scores for comments that differ only in their use of identity terms. To our knowledge, we are the first group to apply tools from Domain Generalization to a fair toxicity classification task.

\section{Experiments}
We evaluate the generalization and fairness properties of IRM\footnote{We implement IRMv1 as specified in Equation \ref{eq:irmv1}, but for brevity refer to this method simply as IRM.} on a variant of a toxic comment detection task. We construct training domains by inducing a positive but spurious correlation between the comment being about a particular demographic identity (e.g., Muslims) and comment toxicity across two training environments, then reverse this correlation strength in the OOD (i.e., ``test'') environment. In this setup, we expect that ERM will predict based on the spurious correlation in the training environments and consequently perform poorly on the OOD environment. We investigate the capacity of IRM to mitigate this failure by learning an invariant predictor robust to this spurious correlation. 

The environment partitions used in these experiments have been constructed in order to control the balance of labels and strength of spurious correlation in each environment. We employ this controlled setup as a proof-of-concept to investigate whether IRM is capable of enabling better generalization in this sort of domain generalization task. However, in practice more ``natural'' environments with a similar correlation structure to our synthetic setup may be found. For instance, analysis of different sub-forums within massive online forums (e.g. Reddit) demonstrates that popular discussion topics along with opinions about these topics can vary dramatically across different sub-forums \citep{redditmining, saleem2017web}. These changes in topics and opinions might indicate the presence of a different degree of racism or sexism in each sub-forum, which in our Domain Generalization framework might correspond to varying strengths of a correlation between the associated racial/sexual identities and toxicity. If each subforum were treated like a different training environment, then IRM may be able to eliminate information about race and sex from its predictors because these features are only spuriously associated with toxicity. This would allow IRM to make fair predictions about comments from new out-of-distributions subforums. Our experiments can be understood as a first step towards evaluating the merit of domain generalization methods like IRM in such settings.

\subsection{Data and problem setup}
We evaluate our model on the Civil Comments dataset,\footnote{\url{https://www.tensorflow.org/datasets/catalog/civil_comments}}
which contains about 2 million comments scraped from the comments section of online news articles that have been manually annotated for toxicity. We use the subset of about 450,000 comments that contain manual annotations indicating whether each comment contains information relevant to thirty-two different demographic groups that could constitute protected group membership in fairness applications (for instance, whether the comment is about a particular race or gender). 

We consider the general task of toxicity classification: predicting the binary toxicity label associated with each comment. Given the often discriminatory nature of online discourse, some demographic identities are more strongly correlated with comment toxicity than others. For instance, about 28\% of ``black'' comments are labelled as toxic compared to an average overall comment toxicity of about 11\% \footnote{See Appendix \ref{sec:exp_details} for details}. An ERM classifier trained on all data might learn these demographic identities as predictors of toxicity, and consequently predict demographic-associated comments as toxic more frequently than non-demographic associated comments, even when both comments have the same true value of toxicity. In our experiments, we consider the following demographic identities -- Black, Muslim, LGBTQ and NeuroDiverse -- to be ``sensitive attributes'': that is, protected groups against which our classifier should not exhibit discriminatory behaviour. See Appendix \ref{sec:exp_details} for further details.

We construct a OOD toxicity classification task in which there are two training environments with strong but spurious correlation between the sensitive attribute and toxicity and one test environment in which this correlation is reversed. All three environments are constrained to be of equal size. Half the comments in each environment are non-toxic ($Y=0$) and half are toxic ($Y=1$). Likewise, half of the comments in each environment are about the sensitive demographic ($Z=1$) and half are not about the sensitive demographic ($Z=0$). Given these constraints, the strength of the spurious correlation in each environment is controlled by modifying the label-switching probability $p^e$. This translates to the following constraints on the distribution of the sensitive attribute given the toxicity label: 

\begin{equation}
    {P(Z=z|Y=1-z) = p^e \ \forall \ z \in \{0, 1\}}
\end{equation}
\begin{equation}
    {P(Z=z|Y=z) = (1 - p^e) \ \forall \ z \in \{0, 1\}}
\end{equation}

The value of $p^e$ varies across the different environments in order to induce the appropriate structure of spurious correlations across the training and testing environments. In the first training environment $p^e = 0.2$; in the second training environment $p^e = 0.1$; in the test environment $p^e = 0.9$. In selecting this configuration of training/testing environments and corresponding values for $p_e$, we follow the precedent of the experiments in \citep{arjovsky2019invariant, choe2019empirical} which also both use only two training environments and these $p_e$ values.

After pre-processing raw text using a standard NLP pipeline, we convert each comment into a sentence embedding representing its semantic content using a pre-trained Sentence-BERT model \citep{reimers2019sentencebert} with base NLI mean tokens (we refer to these embeddings as comment embeddings). 
We separate each of our two training environments into an 80\% training and 20\% validation set. Training environment results are reported on the aggregate of the validation sets of both these environments, which we refer to as the Inside-Of-Distribtuion (InD) set. Testing environment results are reported on all testing environment data, which we refer to as the Out-Of-Distribution (OOD) set.

We train two different logistic regression models with L2 regularization, an ERM model and IRM model, using BCE loss and SGD with Adam \citep{kingma2017adam}. The only 
difference between the two models is that the IRM model is optimized with respect to Equation \ref{eq:irmv1}, whereas the ERM model is optimized to minimize aggregate training loss. See Appendix \ref{sec:exp_details} for training details. 

\subsection{Results}

\paragraph{IRM outperforms ERM on out-of-distribution toxicity classification}  \label{ssec:r1}
\begin{table}[h]
\centering
\begin{tabular}{l|cc|cc}
\toprule
{} & \multicolumn{2}{c|}{In distribution} & \multicolumn{2}{c}{Out of distribution} \\
{}SensAttr &           ERM &            IRM &            ERM &            IRM \\
\midrule
Black    &  85.3  $\pm$   1.0 &  80.6  $\pm$   1.7 &  48.2  $\pm$   1.3 &  58.1  $\pm$   2.7 \\
Muslim   &  84.7  $\pm$   0.7 &  81.1  $\pm$   1.3 &  48.6  $\pm$   0.8 &  57.0  $\pm$   2.3 \\
LGBTQ    &  84.4  $\pm$   0.6 &  78.7  $\pm$   4.0 &  55.7  $\pm$   1.0 &  60.5  $\pm$   3.3 \\
NeuroDiv &  83.0  $\pm$   1.1 &  80.3  $\pm$   1.4 &  63.0  $\pm$   1.0 &  62.5  $\pm$   2.2 \\
\bottomrule
\end{tabular}

\caption{We measure the performance of IRM and ERM on an internet comment toxicity classification task in which mention of particular demographic identities are correlated with toxicity during training, but anti-correlated during testing. The average accuracies (in percentages, with standard deviations over 10 runs) of both methods on data from the training (InD) and testing (OOD) environments are reported. We find that IRM outperforms ERM on the OOD set while maintaining good InD performance for all sensitive attributes except ``NeuroDiv''.}
\label{tab:irm_base_performance}
\end{table}

In Table \ref{tab:irm_base_performance}, we report the mean accuracies for each method on both the InD and OOD sets. We find that for all considered sensitive attributes (with the exception of 'NeuroDiverse'), IRM outperforms ERM on the OOD test set by a difference of 5-10\%, while maintaining comparable performance (within 3-6\%) to ERM on the InD sets. This suggests that IRM is more robust to the spurious correlation than ERM. However, note that IRM's performance improvement is relatively small compared to the maximum possible improvement that IRM could have achieved in this scenario. A worst-case ERM model that exclusively uses the sensitive attribute to predict toxicity could only achieve a maximum of about 10\% accuracy on the OOD set, and we estimate that a best-case IRM model that is completely invariant to the spurious correlation would achieve a maximum accuracy of 80\% (See Appendix \ref{sec:hubs} for details). IRM's failure to achieve the magnitude of performance benefit reported on the toy domain generalization tasks explored in \citep{arjovsky2019invariant, choe2019empirical} may be because the sensitive attribute is more difficult to infer from the data in our experiments, which forces ERM to partially rely on other invariant features of the data for prediction.

\begin{table}[h]
    \centering
    \begin{tabular}{l|cc|cc}
\toprule
{} & \multicolumn{2}{c|}{$\Delta_{EO}$} & \multicolumn{2}{c}{$\Delta_{DP}$} \\
{SensAttr} &            ERM &             IRM &            ERM &            IRM \\
\midrule
Black    &  49.6  $\pm$   3.1 &  29.8  $\pm$   5.3 &  10.3  $\pm$   1.1 &   2.4  $\pm$   2.0 \\
Muslim   &  48.8  $\pm$   1.4 &   32.5  $\pm$   4.8 &  9.8  $\pm$   0.7 &   2.0  $\pm$   1.7 \\
LGBTQ    &  38.2  $\pm$   1.7 &  23.6  $\pm$   6.3 &   2.3  $\pm$   1.0 &   4.3  $\pm$   3.2 \\
NeuroDiv &  26.9  $\pm$   2.3 &   25.0  $\pm$   3.8 &  5.5  $\pm$   0.9 &  5.4  $\pm$   2.3 \\
\bottomrule
\end{tabular}



    \caption{$\Delta_{EO}$ and $\Delta_{DP}$ of both ERM and IRM on the OOD set from the experiments of Table \ref{tab:irm_base_performance}. $\Delta_{DP}$ and $\Delta_{EO}$, respectively, are measures of how much the model violates Demographic Parity and  Equalized Odds that range between 0\% and 100\%. IRM consistently makes fairer predictions than ERM with respect to both fairness definitions. Average values reported with standard deviation over 10 runs.}
    \label{tab:irm_base_fairness_performance}
\end{table}

Table \ref{tab:irm_base_fairness_performance} reports the performance of ERM and IRM on the OOD set with respect to the Demographic Parity and Equalized Odds fairness definitions (Equations \ref{eqn:dp} and \ref{eqn:eo}). We find that IRM yields substantially fairer predictions for both definitions on all considered sensitive attributes (with the exception of Demographic Parity for ``LGBTQ''). These results are expected given the spurious correlation's reversal between the InD and OOD sets and our hypothesis that IRM learns invariant predictors. Since it is more common for demographic identity comments to be non-toxic in the OOD set, ERM -- which is making predictions assuming the opposite correlation between sensitive attribute and toxicity -- will mis-classify demographic identity comments as toxic and regular comments as non-toxic at a higher rate than IRM, whose predictions are not affected by the spurious correlation. This insight both further emphasizes the connection between the fairness and robustness of machine learning classifiers and suggests that fairness and accuracy may be aligned in some cases of distribution shift. An interesting avenue for future work could leverage recent work about the fairness-accuracy tradeoff \citep{NIPS2019_9082, NIPS2018_7625} in order to better understand the conditions under which increasing fairness through robustness also increases accuracy. 

\paragraph{Explaining the performance improvements of IRM} \label{ssec:r2}
\begin{table}[h]
    \centering
    \begin{tabular}{l|cc|cc}
\toprule
{} & \multicolumn{2}{c|}{ERM} & \multicolumn{2}{c}{IRM} \\
SA{} &   non-explicit &       explicit &   non-explicit &       explicit \\
\midrule
Black    &  48.2  $\pm$   1.3 &  38.8  $\pm$   1.4 &  58.1  $\pm$   2.7 &  56.0  $\pm$   3.4 \\
Muslim   &  48.6  $\pm$   0.8 &  40.2  $\pm$   0.9 &  57.0  $\pm$   2.3 &  55.5  $\pm$   4.1 \\
LGBTQ    &  55.7  $\pm$   1.0 &  41.3  $\pm$   1.4 &  60.5  $\pm$   3.3 &  58.4  $\pm$   3.5 \\
NeuroDiv &  63.0  $\pm$   1.0 &  47.0  $\pm$   1.3 &  62.5  $\pm$   2.2 &  56.8  $\pm$   2.1 \\
\bottomrule
\end{tabular}

    \caption{Extending the results of Table \ref{tab:irm_base_performance}, we report average accuracies for ERM and IRM on the testing (OOD) environment both with and without explicitly including the sensitive attribute in the input data representation. We find that including this information reduces accuracy by a larger margin for ERM than for IRM, suggesting that IRM's predictors are more invariant to the sensitive attribute's spurious correlation with toxicity in the training environments.}
    \label{tab:sa_comparison}
\end{table}

Having established that IRM better generalizes to the OOD set, an important question concerns the source of this performance disparity. The explanation suggested in the analogous domain generalization experiments of \citep{arjovsky2019invariant,choe2019empirical} is that because IRM's predictors are encouraged to be invariant to spurious correlations with the sensitive attribute whereas ERM's predictors are simply encouraged to minimize risk across the InD environments, IRM's predictors are robust to the flipped spurious correlation in the OOD set. However, this explanation is not obviously correct; as discussed above, the observed performance disparity between IRM and ERM is much lower than it's theoretical maximum as given by this explanation. 

We test whether the invariance of IRM's predictors is responsible for its increased performance by concatenating a binary sensitive attribute indicator variable to each comment's embedding vector in both the InD and OOD environments. While in our original experiments the models must infer the noisy sensitive attribute indirectly from the data, in this modified version the models are directly and explicitly provided with sensitive attribute information. If the performance disparity between ERM and IRM is due to IRM's invariance with respect sensitive attribute, then after strengthening its ``signal'' in the data ERM should perform more poorly on the OOD set while IRM's performance should remain constant.  

The results of these experiments  are reported in Table \ref{tab:sa_comparison}. We observe that when the sensitive attribute is explicitly included in the input representations, ERM's accuracy drops by an average of about 10-15\% whereas IRM's accuracy only drops by an average of about 2\%.
This disparity in performance degradation suggests that IRM's predictors are more, although not fully, invariant to information about the sensitive attribute than ERM's. This confirms our hypothesis that this IRM's invariance is responsible for the difference in the two methods' performances.

\paragraph{What invariant features are IRM learning?} \label{ssec:r3}

An important interpretability-focused question concerns precisely what invariant features of the text IRM learns to use for prediction. We hypothesize that this key feature is the input comment's length. We replicate the experiments from section \ref{ssec:r1} using two new types of comment embeddings derived from context-free FastText word embeddings \cite{mikolov2018advances}. The ``EmbedSum'' method generates an embedding for whole the comment by summing the embeddings of all words in the comment; conversely, the ``EmbedMean'' method takes the average of these embeddings. The embedding generated for a given comment by EmbedMean is just a scalar multiple of embedding generated by EmbedSum. Therefore, both should represent the same semantic content since their cosine similarity, a commonly used measure of semantic similarity \cite{vor-der-bruck-pouly-2019-text}, is 1. A key difference between embeddings is that EmbedSum represents information about the comment's length whereas EmbedMean does not: on average longer comments will have higher magnitude embeddings with EmbedSum since more word embeddings are being added together. Thus, any difference in performance between these two embedding methods likely result from their different abilities to represent comment length.  

We report the mean accuracies (averaged over all sensitive attributes) of ERM and IRM  for both comment embedding methods (with the sensitive attribute expliclty concatenated) in Table \ref{tab:we_compare_sa}. While EmbedSum yields similar performance to the results computed using SBERT embeddings in Table \ref{tab:sa_comparison}, EmbedMean's performance suggests a catastrophic failure of the model. Firstly ERM, despite achieving 85\% accuracy on the InD set, only achieves 12\% accuracy on the OOD set, which is close to the 10\% expected performance of a worst-case model that exclusively uses the spurious correlation for prediction.  Secondly, IRM achieves random accuracy (50\%) on both the InD and OOD sets which suggests that there are no invariant features of the text encoded in EmbedMean that IRM could have learned. These findings suggest that only EmbedSum encodes the invariant feature of the text responsible for IRM's performance benefit in Tables \ref{tab:irm_base_performance} and \ref{tab:irm_base_fairness_performance}. The best candidate for such an invariant feature is comment length.

\begin{table}[]
    \centering
    \begin{tabular}{l|cc|cc}
\toprule
{} & \multicolumn{2}{c|}{In Distribution} & \multicolumn{2}{c}{Out of Distribution} \\
word\_encoding &            ERM &            IRM &            ERM &            IRM \\
\midrule
EmbedSum      &  84.5  $\pm$   0.8 &  80.0  $\pm$   1.2 &  56.5  $\pm$   3.0 &  62.3  $\pm$   2.4 \\
EmbedMean     &  85.1  $\pm$   0.8 &  50.0  $\pm$   1.5 &  12.0  $\pm$   0.5 &  50.0  $\pm$   0.0 \\
\bottomrule
\end{tabular}

    \caption{We replicate the experiments of Table \ref{tab:irm_base_performance} (averaged over all demographic identities) but instead embed comment text using the sum or mean of FastText word embeddings for all words in the comment. Both ERM and IRM generalize to the OOD set using the EmbedSum method but do not when using the EmbedMean method, suggesting that the comment length is an invariant predictor of toxicity.}
    \label{tab:we_compare_sa}
\end{table}

These findings call for caution and discretion when applying IRM to real-world fairness problems. Although comment length happens to be a somewhat robust predictor of toxicity for comments in the Jigsaw dataset, it may not be as effective for comments that come from other domains. For instance, our IRM model would likely be less effective at predicting the toxicity on twitter comments since their lengths are tightly constrained by the platform's design. This speaks to the importance of using a selection of training environments which are diverse enough that they only share invariances that are intrinsically characteristic of the phenomenon being predicted. A similar word of warning applies to assigning causal interpretations the predictors generated by IRM and other Domain Generalization algorithms. While comment length may be an invariant predictor of toxicity, it is clearly not a cause of toxicity. An interesting avenue of future work concerns the conditions under which environments are sufficiently diverse to only include the invariances of interest.

\section{Conclusion}
In this work, we applied IRM to a toxicity classification task in order to demonstrate that Domain Generalization can serve as an important framework for building fair machine learning classifiers. Our findings show that IRM outperforms ERM with respect to both generalization accuracy and group fairness by learning invariant but likely non-causal predictors of toxicity. We hope that these results are first steps for future explorations of the relationship between robustness and fairness in machine learning.

{\small
\bibliographystyle{plainnat}
\bibliography{refs}  

\begin{thebibliography}{40}
\providecommand{\natexlab}[1]{#1}
\providecommand{\url}[1]{\texttt{#1}}
\expandafter\ifx\csname urlstyle\endcsname\relax
  \providecommand{\doi}[1]{doi: #1}\else
  \providecommand{\doi}{doi: \begingroup \urlstyle{rm}\Url}\fi

\bibitem[jig()]{jigsawmedium}
Unintended bias and names of frequently targeted groups.
\newblock
  \url{https://medium.com/the-false-positive/unintended-bias-and-names-of-frequently-targeted-groups-8e0b81f80a23}.
\newblock Accessed: 2020-19-08.

\bibitem[per()]{perspective}
Perspective api.
\newblock \url{https://www.perspectiveapi.com/#/home}.
\newblock Accessed: 2020-11-10.

\bibitem[Arjovsky et~al.(2019)Arjovsky, Bottou, Gulrajani, and
  Lopez-Paz]{arjovsky2019invariant}
Martin Arjovsky, L{\'e}on Bottou, Ishaan Gulrajani, and David Lopez-Paz.
\newblock Invariant risk minimization.
\newblock \emph{arXiv preprint arXiv:1907.02893}, 2019.

\bibitem[Barocas and Selbst(2016)]{barocas2016big}
Solon Barocas and Andrew~D Selbst.
\newblock Big data's disparate impact.
\newblock \emph{Calif. L. Rev.}, 104:\penalty0 671, 2016.

\bibitem[Bolukbasi et~al.(2016)Bolukbasi, Chang, Zou, Saligrama, and
  Kalai]{bolukbasi2016man}
Tolga Bolukbasi, Kai-Wei Chang, James Zou, Venkatesh Saligrama, and Adam Kalai.
\newblock Man is to computer programmer as woman is to homemaker? debiasing
  word embeddings, 2016.

\bibitem[Borkan et~al.(2019)Borkan, Dixon, Sorensen, Thain, and
  Vasserman]{borkan2019nuanced}
Daniel Borkan, Lucas Dixon, Jeffrey Sorensen, Nithum Thain, and Lucy Vasserman.
\newblock Nuanced metrics for measuring unintended bias with real data for text
  classification, 2019.

\bibitem[Brown et~al.(2020)Brown, Mann, Ryder, Subbiah, Kaplan, Dhariwal,
  Neelakantan, Shyam, Sastry, Askell, Agarwal, Herbert-Voss, Krueger, Henighan,
  Child, Ramesh, Ziegler, Wu, Winter, Hesse, Chen, Sigler, Litwin, Gray, Chess,
  Clark, Berner, McCandlish, Radford, Sutskever, and Amodei]{brown2020language}
Tom~B. Brown, Benjamin Mann, Nick Ryder, Melanie Subbiah, Jared Kaplan,
  Prafulla Dhariwal, Arvind Neelakantan, Pranav Shyam, Girish Sastry, Amanda
  Askell, Sandhini Agarwal, Ariel Herbert-Voss, Gretchen Krueger, Tom Henighan,
  Rewon Child, Aditya Ramesh, Daniel~M. Ziegler, Jeffrey Wu, Clemens Winter,
  Christopher Hesse, Mark Chen, Eric Sigler, Mateusz Litwin, Scott Gray,
  Benjamin Chess, Jack Clark, Christopher Berner, Sam McCandlish, Alec Radford,
  Ilya Sutskever, and Dario Amodei.
\newblock Language models are few-shot learners, 2020.

\bibitem[Buolamwini and Gebru(2018)]{buolamwini18a}
Joy Buolamwini and Timnit Gebru.
\newblock Gender shades: Intersectional accuracy disparities in commercial
  gender classification.
\newblock volume~81 of \emph{Proceedings of Machine Learning Research}, pages
  77--91, New York, NY, USA, 23--24 Feb 2018. PMLR.
\newblock URL \url{http://proceedings.mlr.press/v81/buolamwini18a.html}.

\bibitem[Cartwright(2003)]{cartwright2003two}
Nancy Cartwright.
\newblock Two theorems on invariance and causality.
\newblock \emph{Philosophy of Science}, 70, 2003.

\bibitem[Choe et~al.(2019)Choe, Ham, and Park]{choe2019empirical}
Yo~Joong Choe, Jiyeon Ham, and Kyubyong Park.
\newblock An empirical study of invariant risk minimization.
\newblock \emph{arXiv preprint arXiv:2004.05007}, 2019.

\bibitem[Dwork et~al.(2011)Dwork, Hardt, Pitassi, Reingold, and
  Zemel]{dwork2011fairness}
Cynthia Dwork, Moritz Hardt, Toniann Pitassi, Omer Reingold, and Rich Zemel.
\newblock Fairness through awareness, 2011.

\bibitem[Ganin et~al.(2016)Ganin, Ustinova, Ajakan, Germain, Larochelle,
  Laviolette, Marchand, and Lempitsky]{ganin2016domainadversarial}
Yaroslav Ganin, Evgeniya Ustinova, Hana Ajakan, Pascal Germain, Hugo
  Larochelle, François Laviolette, Mario Marchand, and Victor Lempitsky.
\newblock Domain-adversarial training of neural networks, 2016.

\bibitem[Garg et~al.(2019)Garg, Perot, Limtiaco, Taly, Chi, and
  Beutel]{garg2019counterfactual}
Sahaj Garg, Vincent Perot, Nicole Limtiaco, Ankur Taly, Ed~H Chi, and Alex
  Beutel.
\newblock Counterfactual fairness in text classification through robustness.
\newblock In \emph{Proceedings of the 2019 AAAI/ACM Conference on AI, Ethics,
  and Society}, pages 219--226, 2019.

\bibitem[Geirhos et~al.(2020)Geirhos, Jacobsen, Michaelis, Zemel, Brendel,
  Bethge, and Wichmann]{geirhos2020shortcut}
Robert Geirhos, Jörn-Henrik Jacobsen, Claudio Michaelis, Richard Zemel,
  Wieland Brendel, Matthias Bethge, and Felix~A. Wichmann.
\newblock Shortcut learning in deep neural networks, 2020.

\bibitem[Goodfellow et~al.(2015)Goodfellow, Shlens, and
  Szegedy]{goodfellow2015explaining}
Ian~J. Goodfellow, Jonathon Shlens, and Christian Szegedy.
\newblock Explaining and harnessing adversarial examples, 2015.

\bibitem[Gulrajani and Lopez-Paz(2020)]{gulrajani2020search}
Ishaan Gulrajani and David Lopez-Paz.
\newblock In search of lost domain generalization, 2020.

\bibitem[Hardt et~al.(2016)Hardt, Price, and Srebro]{hardt2016equality}
Moritz Hardt, Eric Price, and Nathan Srebro.
\newblock Equality of opportunity in supervised learning, 2016.

\bibitem[Heinze-Deml et~al.(2018)Heinze-Deml, Peters, and
  Meinshausen]{heinzedeml2018invariant}
Christina Heinze-Deml, Jonas Peters, and Nicolai Meinshausen.
\newblock Invariant causal prediction for nonlinear models, 2018.

\bibitem[Khan and Golab(2020)]{redditmining}
Abeer Khan and Lukasz Golab.
\newblock Reddit mining to understand gendered movements.
\newblock In \emph{Proceedings of the 23rd International Conference on
  Extending Database Technology}, 2020.

\bibitem[Kingma and Ba(2017)]{kingma2017adam}
Diederik~P. Kingma and Jimmy Ba.
\newblock Adam: A method for stochastic optimization, 2017.

\bibitem[Kiritchenko and Mohammad(2018)]{kiritchenko2018examining}
Svetlana Kiritchenko and Saif~M. Mohammad.
\newblock Examining gender and race bias in two hundred sentiment analysis
  systems, 2018.

\bibitem[Kusner et~al.(2018)Kusner, Loftus, Russell, and
  Silva]{kusner2018counterfactual}
Matt~J. Kusner, Joshua~R. Loftus, Chris Russell, and Ricardo Silva.
\newblock Counterfactual fairness, 2018.

\bibitem[Li et~al.(2018)Li, Gong, Tian, Liu, and Tao]{li2018domain}
Ya~Li, Mingming Gong, Xinmei Tian, Tongliang Liu, and Dacheng Tao.
\newblock Domain generalization via conditional invariant representation, 2018.

\bibitem[Mansour and Schain(2014)]{mansour2019robust}
Yishay Mansour and Mariano Schain.
\newblock Robust domain adaptation.
\newblock In \emph{Annals of Mathematics and Artificial Intellignece}, pages
  365--380, August 2014.
\newblock \doi{10.1007/s10472-013-9391-5}.

\bibitem[Mikolov et~al.(2018)Mikolov, Grave, Bojanowski, Puhrsch, and
  Joulin]{mikolov2018advances}
Tomas Mikolov, Edouard Grave, Piotr Bojanowski, Christian Puhrsch, and Armand
  Joulin.
\newblock Advances in pre-training distributed word representations.
\newblock In \emph{Proceedings of the International Conference on Language
  Resources and Evaluation (LREC 2018)}, 2018.

\bibitem[Mitchell et~al.(2019)Mitchell, Wu, Zaldivar, Barnes, Vasserman,
  Hutchinson, Spitzer, Raji, and Gebru]{mitchell2019model}
Margaret Mitchell, Simone Wu, Andrew Zaldivar, Parker Barnes, Lucy Vasserman,
  Ben Hutchinson, Elena Spitzer, Inioluwa~Deborah Raji, and Timnit Gebru.
\newblock Model cards for model reporting.
\newblock In \emph{Proceedings of the conference on fairness, accountability,
  and transparency}, pages 220--229, 2019.

\bibitem[Mitchell(2000)]{mitchell2000dimensions}
Sandra Mitchell.
\newblock Dimensions of scientific law.
\newblock \emph{Philosophy of Science}, 67, 2000.

\bibitem[Peters et~al.(2015)Peters, Bühlmann, and
  Meinshausen]{peters2015causal}
Jonas Peters, Peter Bühlmann, and Nicolai Meinshausen.
\newblock Causal inference using invariant prediction: identification and
  confidence intervals, 2015.

\bibitem[Ravfogel et~al.(2020)Ravfogel, Elazar, Gonen, Twiton, and
  Goldberg]{ravfogel2020null}
Shauli Ravfogel, Yanai Elazar, Hila Gonen, Michael Twiton, and Yoav Goldberg.
\newblock Null it out: Guarding protected attributes by iterative nullspace
  projection, 2020.

\bibitem[Reimers and Gurevych(2019)]{reimers2019sentencebert}
Nils Reimers and Iryna Gurevych.
\newblock Sentence-bert: Sentence embeddings using siamese bert-networks, 2019.

\bibitem[Saleem et~al.(2017)Saleem, Dillon, Benesch, and Ruths]{saleem2017web}
Haji~Mohammad Saleem, Kelly~P Dillon, Susan Benesch, and Derek Ruths.
\newblock A web of hate: Tackling hateful speech in online social spaces, 2017.

\bibitem[Sap et~al.(2019)Sap, Card, Gabriel, Choi, and
  Smith]{sap-etal-2019-risk}
Maarten Sap, Dallas Card, Saadia Gabriel, Yejin Choi, and Noah~A. Smith.
\newblock The risk of racial bias in hate speech detection.
\newblock In \emph{Proceedings of the 57th Annual Meeting of the Association
  for Computational Linguistics}, pages 1668--1678, Florence, Italy, July 2019.
  Association for Computational Linguistics.
\newblock \doi{10.18653/v1/P19-1163}.
\newblock URL \url{https://www.aclweb.org/anthology/P19-1163}.

\bibitem[Sweeney and Najafian(2020)]{sweeny2020reducing}
Chris Sweeney and Maryam Najafian.
\newblock Reducing sentiment polarity for demographic attributes in word
  embeddings using adversarial learning.
\newblock In \emph{Proceedings of the 2020 Conference on Fairness,
  Accountability, and Transparency}, 2020.

\bibitem[Tan and Celis(2019)]{tan2019assessing}
Yi~Chern Tan and L.~Elisa Celis.
\newblock Assessing social and intersectional biases in contextualized word
  representations, 2019.

\bibitem[Vaidya et~al.(2019)Vaidya, Mai, and Ning]{vaidya2019empirical}
Ameya Vaidya, Feng Mai, and Yue Ning.
\newblock Empirical analysis of multi-task learning for reducing model bias in
  toxic comment detection.
\newblock \emph{arXiv preprint arXiv:1909.09758}, 2019.

\bibitem[vor~der Br{\"u}ck and Pouly(2019)]{vor-der-bruck-pouly-2019-text}
Tim vor~der Br{\"u}ck and Marc Pouly.
\newblock Text similarity estimation based on word embeddings and matrix norms
  for targeted marketing.
\newblock In \emph{Proceedings of the 2019 Conference of the North {A}merican
  Chapter of the Association for Computational Linguistics: Human Language
  Technologies, Volume 1 (Long and Short Papers)}, pages 1827--1836,
  Minneapolis, Minnesota, June 2019. Association for Computational Linguistics.
\newblock \doi{10.18653/v1/N19-1181}.
\newblock URL \url{https://www.aclweb.org/anthology/N19-1181}.

\bibitem[Walker et~al.(2012)Walker, Tree, Anand, Abbott, and
  King]{walker-etal-2012-corpus}
Marilyn Walker, Jean~Fox Tree, Pranav Anand, Rob Abbott, and Joseph King.
\newblock A corpus for research on deliberation and debate.
\newblock In \emph{Proceedings of the Eighth International Conference on
  Language Resources and Evaluation ({LREC}'12)}, pages 812--817, Istanbul,
  Turkey, May 2012. European Language Resources Association (ELRA).
\newblock URL
  \url{http://www.lrec-conf.org/proceedings/lrec2012/pdf/1078_Paper.pdf}.

\bibitem[Wick et~al.(2019)Wick, panda, and Tristan]{NIPS2019_9082}
Michael Wick, swetasudha panda, and Jean-Baptiste Tristan.
\newblock Unlocking fairness: a trade-off revisited.
\newblock In H.~Wallach, H.~Larochelle, A.~Beygelzimer, F.~d~Alch\'{e}-Buc,
  E.~Fox, and R.~Garnett, editors, \emph{Advances in Neural Information
  Processing Systems 32}, pages 8783--8792. Curran Associates, Inc., 2019.
\newblock URL
  \url{http://papers.nips.cc/paper/9082-unlocking-fairness-a-trade-off-revisited.pdf}.

\bibitem[Zhang and Bareinboim(2018)]{NIPS2018_7625}
Junzhe Zhang and Elias Bareinboim.
\newblock Equality of opportunity in classification: A causal approach.
\newblock In S.~Bengio, H.~Wallach, H.~Larochelle, K.~Grauman, N.~Cesa-Bianchi,
  and R.~Garnett, editors, \emph{Advances in Neural Information Processing
  Systems 31}, pages 3671--3681. Curran Associates, Inc., 2018.
\newblock URL
  \url{http://papers.nips.cc/paper/7625-equality-of-opportunity-in-classification-a-causal-approach.pdf}.

\bibitem[Zhao et~al.(2020)Zhao, Li, Reed, Xu, and Keutzer]{zhao2020multisource}
Sicheng Zhao, Bo~Li, Colorado Reed, Pengfei Xu, and Kurt Keutzer.
\newblock Multi-source domain adaptation in the deep learning era: A systematic
  survey, 2020.

\end{thebibliography}
}

\appendix

\section{Experimental details}\label{sec:exp_details}
\paragraph{Relating our experiments to CMNIST}
Our experimental design is inspired by the Colored MNIST (CMNIST) experiments from \citep{arjovsky2019invariant}: a classification task in which different colours are assigned to different digits present in MNIST images and these images are split into different environments such that colour is spuriously correlated with outcome digit. However, our setup presents a more realistic assessment of IRM's performance than the CMNIST experiments for the following two reasons:

\begin{itemize}
    \item In the CMNIST experiments the spurious covariate (colour) is easily inferable from image data using RGB channel information. In our experiments the spurious covariate (demographic identity) is not as directly inferable from the data. Annotators were asked to assign identities to comments based on their interpretations of the comment content, which may be presented using a wide variety of different word choices and sentence structures \citep{borkan2019nuanced}. It consequently may be more difficult for models in our experiments to construct predictors using information about the sensitive attribute. 
    \item There is no noise on the ground-truth labels in CMNIST experiments (digit shape), whereas the labels in our experiments (toxicity) are known to be noisy \citep{walker-etal-2012-corpus}. Although a 25\% chance of label corruption is also induced for the CMNIST experiments in \citep{arjovsky2019invariant}, this noise is applied to all samples. This is likely not the case in our experiments; toxicity annotations of social media text have been found to be biased against sensitive demographics such that it is more likely for more of their comments to be toxic \citep{sap-etal-2019-risk}. This differently distributed noise across different sensitive attributes makes it less certain that toxicity is genuinely an invariant feature, increasing the difficulty of the problem. 
\end{itemize}

\paragraph{Model selection} 
Hyperparameter selection presents a unique challenge for domain generalization problems. Since the model is not permitted to have access to the test environment during training,  ``validation'' data that is distributed identically to the test  environment is not available for hyperparameter tuning. We use the the ``oracle selection strategy'' suggested by \citep{gulrajani2020search} in which hyperparameters are tuned directly on the OOD set, but only a limited number of possible hyperparameter settings are considered as possible candidates. We hope that fixing the number of considered settings in advance will adequately limit access to the test distribution and, since this selection strategy is also employed in \citep{arjovsky2019invariant}, we must use this strategy for fair comparison with these results. 

For both the ERM and IRM models we consider a selection of 50 different hyperparameter settings randomly chosen from a uniform distribution over all parameters with plausible upper and lower bounds. Of these 50 candidate settings we remove from consideration those whose corresponding model overfits to the training environments, defined as those for which the absolute difference between InD set loss and training data loss is greater than 0.05. This is particularly a problem for ERM models, which can perform extremely well by over-fitting to the point where it gives random performance on the OOD set, which is much better than its OOD set performance if it's learned the spurious correlation. Then, of the remaining settings, we select the one with the highest aggregated accuracy on the InD and OOD sets, averaged over 5 trials. 

The lower and upper bounds of each hyperparameter are given below
\begin{itemize}
    \item learning rate: (0.0001, 0.01)
    \item number of iterations: (10000, 100000)
    \item l2 regularization: (0.00001, 0.1)
    \item IRM invariance penalty: (100, 1000000)
    \item penalty anneal iterations: (50, 250)
\end{itemize}

\paragraph{Model architecture}
We use a logistic regression with l2 regularization as our baseline toxicity classification model. State-of-the-art performance on many NLP tasks is achieved using much more complex models than logistic regression \citep{brown2020language}, and in principle these kinds of models may also be trained using an IRM invariance penalty. We choose to study logistic regression given that lower-capacity models are more likely than higher-capacity models to absorb strong but spurious correlations in the data. Consequently, they should experience greater performance improvement that higher-capacity models after adding an IRM invariance penalty to the loss function \citep{arjovsky2019invariant}. 

\paragraph{Sensitive attribute details}
The Civil Comments Dataset contains information relevant to thirty-two different demographic groups that could constitute protected group membership in fairness applications. Every comment is associated with an ``group label'' for each of the thirty-two groups, consisting of a score between zero and one indicating the degree to which the comment is about that group. We binarize these labels such that each comment is ``about'' a given group if it's label for that group is grater than zero; this binarization threshold is so low given the lack of positive labels for all demographic groups.

Our experiments evaluate four different ``demographic identities'' as sensitive attributes, where each identity is a cluster of related groups defined in Civil Comments. The groups which constitute each identity are given in Table \ref{tab:sensatt}. We only include comments that are exclusively about one of the four sensitive attributes in our experiments; comments that are about multiple attributes are ignored.
\begin{table}[]
    \centering
    \begin{tabular}{|l|p{100mm}|}
\hline
Sensitive Attribute & Civil Comments Groups                                                                              \\ \hline
Black               & black                                                                                              \\ \hline
Muslim              & muslim                                                                                             \\ \hline
LGBTQ               & homosexual\_gay\_or\_lesbian --  bisexual --  other\_sexual\_orientation --  \newline other\_gender -- transgender \\ \hline
NeuroDiv            & psychiatric\_or\_mental\_illness -- intellectual\_or\_learning\_disability                          \\ \hline
\end{tabular}
    \caption{The Civil Comments defined demographic groups constituting the sensitive attributes in each of our experiments}
    \label{tab:sensatt}
\end{table}

\paragraph{Toxicity labels details}
Each comment in the Civil Comments Dataset is associated with a ``toxicity label'' between zero and one indicating the comment's degree of toxicity - where a toxic comment is defined as ``a rude, disrespectful, or unreasonable comment that is likely to make you leave a discussion'' \citep{perspective}. For all experiments, unless otherwise specified, we binarize these labels such that toxic comments have a toxicity score of 0.5 or greater and non-toxic comments have a score of less than 0.5.

\paragraph{Heuristic upper bound on IRM performance} \label{sec:hubs}
In the CMNIST experiments outlined in \citep{arjovsky2019invariant}, colour (the spurious covariate) is assigned to different labels synthetically. Therefore, it is possible to test the performance of an ``oracle method'' on this domain generalization task that is completely invariant to the spurious correlation by running the experiment without colouring any of the images. This approach is unfortunately not possible in our experiments because the demographic identity to which a given comment refers cannot be trivially ``removed'' from the comment. A comment can be about some demographic identity in many ways; for instance, a comment tagged ``muslim'' could be discussing the political situation in a country with a majority Muslim population, or racistly demanding that immigrants ``go back to their own countries'', or wishing others a happy Eid al-Ftir. It is difficult to see how one could generate versions of these sorts of comments that retain the features which contribute to their toxicity while removing all reference to demographic-identity.

Instead, we estimate this ``oracle model'' as a logistic regression trained on a toxicity-balanced subset of the Jigsaw comments that do not reference any demographic identities. Since the lack of demographic identity information in the comments eliminates any possible effects of spurious correlations between the sensitive attribute and labels, the model's performance represent what IRM's performance on the OOD set would have been if IRM were completely invariant to spurious correlations. We treat the model's performance on this dataset as a heuristic `upper bound'' for IRM's performance in our main experiments. 

We train our oracle model on a toxicity-balanced dataset of about 26000 samples with a 70-15-15 train-validation-test split and use grid search to tune learning rate and l2 regularization. Our best model, with a learning rate of 0.01 and l2 regularization of 0.005, yields a test set accuracy of 80.3\%

\end{document}